\documentclass[preprint,12pt]{elsarticle}

\usepackage{amsmath,amssymb}
\usepackage{graphicx}
\usepackage{booktabs}
\usepackage{hyperref}
\usepackage{threeparttable}

\usepackage{tikz}
\usepackage{subcaption}
\usepackage{xcolor}
\usetikzlibrary{arrows.meta, positioning, decorations.pathreplacing, calc}

\usepackage{enumitem}
\setlist[itemize]{leftmargin=*}
\begin{document}

\begin{frontmatter}


\title{\large{The Effective Depth Paradox: Topology and Trainability in Deep CNNs}}

\author[inst1]{Manfred M. Fischer\corref{cor1}}

\author[inst2]{Joshua Pitts}
\address[inst1]{Vienna University of Economics and Business, Vienna, Austria}
\address[inst2]{Boston University, Boston, USA}
\cortext[cor1]{Corresponding author, Vienna University of Economics and Business, Department of Socioeconomics, Welthandelsplatz 1, 1020 Vienna; email: manfred.fischer@wu.ac.at}
\vspace{-1cm}


\begin{abstract} 
\noindent This paper presents a controlled comparative study of convolutional neural network (CNN) topology and image classification performance across the architectural families VGG, ResNet, and GoogLeNet, evaluated on CIFAR-10 under a unified training protocol. We formalize the distinction between nominal depth ($D_{\mathrm{nom}}$), the physical count of weight-bearing layers, and effective depth ($D_{\mathrm{eff}}$), an operational metric quantifying the expected length of forward information paths, extending the path-ensemble interpretation of residual networks introduced by Veit et al. (2016) into closed-form, pre-training proxies spanning sequential, residual, and multi-branch topologies. We validate this proxy against a gradient-weighted variant computed from observed backpropagation signal. Across eight representative models (VGG-11/13/16/19, ResNet-18/34/50, GoogLeNet), plain VGG-style stacks show early accuracy saturation as $D_{\mathrm{eff}}$ increases, whereas ResNet and GoogLeNet continue to benefit from added depth by keeping $D_{\mathrm{eff}}$ low relative to $D_{\mathrm{nom}}$ — a pattern we term the “Effective Depth Paradox”. A pooled correlation analysis shows both $D_{\mathrm{nom}}$ and $D_{\mathrm{eff}}$ are strongly, significantly associated with accuracy (r = 0.94 and r = 0.93; both p < 0.01); given the small family-clustered sample, this alone cannot cleanly separate the two metrics, so we treat gradient-norm evidence as complementary mechanistic support rather than decisive statistical proof. We conclude that architectural topology, not layer count alone, governs trainability and scaling efficiency in deep CNNs. All claims are scoped to CIFAR-10-scale training of the three families studied; we do not claim validation at ImageNet scale or generalization to modern architectures such as EfficientNet, ConvNeXt, or Vision Transformers, which we identify as necessary future work.

\end{abstract}

\end{frontmatter}





\section{Introduction}
\label{sec:introduction}

\noindent The depth of convolutional neural networks (CNNs) is a major driver of progress in image recognition, significantly improving representational power (LeCun et al. \cite{lecun2015deep}). However, the relationship between depth and performance is increasingly complex. As networks expand, they often face convergence challenges, most notably vanishing and exploding gradients (Hochreiter \cite{hochreiter1998vanishing}). In plain sequential networks, gradient signals can decay exponentially with depth, leading to optimization instability (Glorot and Bengio \cite{glorot2010understanding}). These limitations motivated the development of architectural mechanisms — such as residual connections and multi-branch modules — to stabilize gradient flow.

Recent literature suggests that residual networks can be interpreted as an ensemble of paths of varying lengths (Veit et al. \cite{veit2016residual}), where shortcut connections create "gradient highways" that bypass the vanishing-gradient problem. From the perspective of continuous dynamical systems, deep architectures can also be viewed as discretized gradient flows, in which stable architectures ensure that the signal remains bounded across hundreds of layers (Haber and Ruthotto \cite{haber2017stable}).

While existing scaling laws predominantly focus on parameter count and training data volume, they often overlook the internal topological efficiency that governs the model's actual capacity. The gap between manual architectural design and automated optimization remains wide, as highlighted by Mienye and Swart \cite{Mienye2024Comprehensive}, who argue that multi-path information flow is a more reliable indicator of representational power than raw depth.

Since the landmark success of AlexNet (Krizhevsky et al. \cite{krizhevsky2012imagenet}), three architectures have emerged as canonical representatives of different depth paradigms: VGG networks (Simonyan and Zisserman \cite{simonyan2015vgg}), ResNet (He et al. \cite{he2016resnet}), and GoogLeNet  (Szegedy et al. \cite{szegedy2015googlenet}). Although all three achieve strong performance, they differ fundamentally in how depth is structured and how information and gradients propagate. Empirical evaluations often report performance metrics based solely on nominal depth ($D_{\mathrm{nom}}$) without accounting for the architectural mechanisms that govern effective depth ($D_{\mathrm{eff}}$). This can lead to misleading conclusions about the benefits of increasing depth — a phenomenon we refer to, following the ensemble-path literature, as the "Effective Depth Paradox": within this study's scope, networks lacking skip connections gain comparatively little from additional nominal depth, while networks that decouple $D_{\mathrm{eff}}$ from $D_{\mathrm{nom}}$ continue to benefit.

This paper presents a standardized comparative study of three CNN architectures, focusing on the distinction between nominal depth (the total number of layers) and effective depth (the expected length of information pathways) and its implications for image recognition under a fixed, controlled training protocol. The study analyzes how increasing convolutional depth influences classification accuracy, convergence dynamics, and computational efficiency in VGG, ResNet, and GoogLeNet. By standardizing training protocols, the research provides empirical evidence within the tested regime that the benefits of depth depend on architectural mechanisms that stabilize gradient flow during training. The findings show that while plain networks such as VGG exhibit early accuracy saturation under our protocol, residual and Inception-based architectures leverage additional depth more effectively. The work aims to demonstrate that effective depth is a useful quantity that mediates among nominal depth, optimization stability, and performance in the models tested, not to establish a general law governing all deep learning architectures.

\subsection{Scope, Contribution, and Structure of the Paper}

\noindent To state the contribution precisely: this paper does not claim to introduce the interpretation of the path-averaging of residual networks, which originates from Veit et al. \cite{veit2016residual}. Our contribution is (i) a unifying, closed-form operationalization of that idea that extends cleanly to multi-branch topologies such as Inception modules, which the ensemble interpretation does not by itself cover; (ii) a controlled, cross-family empirical comparison under a single training protocol, designed specifically to isolate topology from confounding implementation choices; and (iii) an explicit cross-check of the resulting structural metric against a gradient-weighted variant computed from observed training dynamics (Appendix B), rather than assuming the structural proxy is adequate. All empirical claims are restricted to the eight models, the CIFAR-10 dataset (with resolution 224 × 224), and the training protocol described in Section \ref{sec:setup}; we return to the question of external validity in Section \ref{sec:limitations}.

The remainder of this paper is organized as follows.  Section \ref{sec:related} reviews the related literature and positions the contribution of the paper explicitly against Veit et al.\cite{veit2016residual}. Section \ref{sec:architectures} provides the formal mathematical framework, detailing the architectural proxies used to calculate $D_{\mathrm{nom}}$ and $D_{\mathrm{eff}}$ for heterogeneous topologies. Section \ref{sec:setup} outlines the unified experimental design and its statistical analysis protocol. Section \ref{sec:results} presents the empirical results, including a dedicated correlation analysis. Section \ref{sec:discussion} discusses the "Effective Depth Paradox" and its (deliberately bounded) implications. Section \ref{sec:conclusions} closes with strengths, an expanded discussion of limitations, and future research directions.

\section{Related Work}
\label{sec:related}

\noindent Although early deep convolutional neural network (DCNN) research focused primarily on raw depth, contemporary perspectives emphasize structural efficiency. Beyond the canonical VGG and ResNet families, the pattern-recognition community has pioneered modular designs that optimize feature extraction. Singh et al. \cite{singh2022context} investigate how architectural modules can enhance feature extraction in DCNNs. The survey by Mienye and Swart \cite{Mienye2024Comprehensive} highlights that multi-path information processing is now a primary driver of representational capacity, moving beyond simple layer stacking. Building on this notion, Khan et al. \cite{Khan2023ASU} demonstrate that integrating such efficient architectures with advanced feature visualizations is important to achieve high performance and interpretability on standardized benchmarks such as CIFAR-10.

\subsection{The Theoretical Benefits and Practical Costs of Depth}

\noindent Early theoretical work established that deeper networks possess expressive power that grows exponentially with the number of layers (Telgarsky \cite{telgarsky2016benefits}; Raghu et al. \cite{raghu2017expressive}). However, in sequential models, this "benefit of depth" is frequently negated by the vanishing-gradient problem. Research in the pattern-recognition literature has highlighted that simply increasing nominal depth often leads to a "shattered" optimization landscape (Glorot and Bengio \cite{glorot2010understanding}). Consequently, modern studies emphasize the need for functional architectures that leverage depth without incurring prohibitive optimization costs.

\subsection{Architectural Solutions and Multi-Branching in Pattern Recognition}

\noindent To bypass the limitations of sequential depth, two major architectural paradigms have emerged: residual learning and multi-branching.

\begin{itemize}	
\item \textbf{Residual Connections}: Identity
 shortcuts transform the network into an ensemble of paths, facilitating stable gradient flow (He et al. \cite{he2016identity}; Srivastava et al. \cite{srivastava2015highway}). Recent work has explored how these connections can be optimized using differentiable learning to produce even more efficient architectures (Guo et al. \cite{guo2022diff}).
 
 \item \textbf{Multi-Branch Modules}: Multi-branching, as popularized by Inception, increases a model's width and multi-scale processing capabilities without increasing the sequential length of the gradient path (Szegedy et al. \cite{szegedy2015googlenet}). Recent contributions have introduced bidirectional parallel multi-branch modules to enhance feature expression across scales, particularly for remote sensing and aerial imagery, see e.g., Li et al. \cite{li2021bidirectional}, which focus on feature pyramid networks for remote sensing.
 \end{itemize}

\subsection{Measuring Network Complexity}
\noindent Standard metrics such as parameter counts or floating-point operations often fail to capture an architecture's true trainability (Guarrasi et al. \cite{guarrasi2022pareto}). This gap has motivated a surge of Neural Architecture Search (NAS) research.

\begin{itemize}
 \item \textbf{NAS and Efficiency}:
 Lightweight NAS methods have been proposed to discover optimal stacked substructures that surpass manually designed models in accuracy while maintaining a low computational footprint (Lin et al. \cite{lin2024lightweight}).
 
\item \textbf{Gradient Flow Stability}:
Researchers have begun using gradient flow analysis based on bottleneck layers to derive a theoretical basis for performance differences between models such as VGG and ResNet.
\end{itemize}


\subsection{Positioning Relative to Veit et al. (2016) and Statement of Contribution}
\label{sec:positioning}

\noindent The central idea underlying effective depth — that a residual network with identity shortcuts implicitly instantiates an ensemble of paths of differing lengths and that its expected path length is therefore shorter than its nominal depth — is the central result of Veit et al. \cite{veit2016residual}. However, that paper does not provide (i) a closed-form proxy usable without exhaustively enumerating paths, (ii) an extension of the same logic to non-residual multi-branch topologies such as Inception modules, or (iii) a cross-family empirical comparison against a shared training protocol. 

The path-averaging idea is computable directly from an architecture's specification, before any training occurs; second, we extend the underlying logic, in a structurally analogous but not identical form, to multi-branch (Inception-style) topologies, for which the ensemble-of-paths interpretation was not originally formulated; and third, we validate the resulting structural proxy empirically against a gradient-weighted variant computed from observed backpropagation signal (\ref{app:gradient_effective_depth}), rather than assuming the uniform path-averaging approximation is adequate. The GoogLeNet branch-averaging formula (Eq. (\ref{eq:eff_googlenet})) is a natural direct extension of the same averaging logic used for residual blocks and is not presented as a conceptually distinct innovation.

Read this way, the paper's role is to operationalize and empirically stress-test an existing theoretical insight across a broader set of topologies than it was originally applied to, and to report honestly on the resulting evidence — including where that evidence is inconclusive at the sample size available (Section \ref{sec: statistical_association}).

\section{Architectures and Depth Definition}
\label{sec:architectures}

\subsection{Network Architectures}
\label{sec:network}

\begin{figure}[t]
\centering
\vspace{-2mm}
\includegraphics[width=\linewidth]{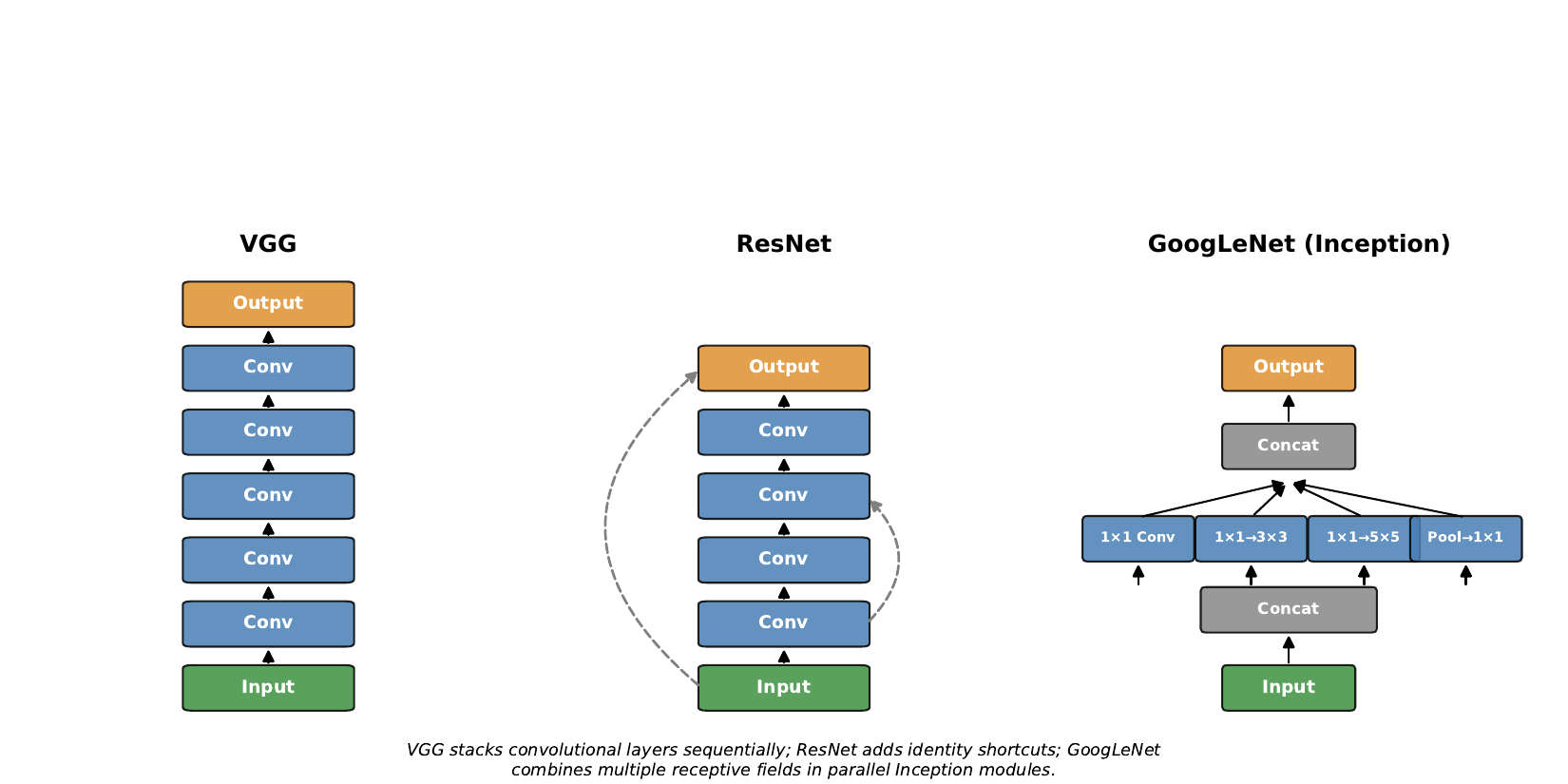}
\caption{\textbf {Schematic overview of the evaluated architectures. VGG employs uniform stacking of convolutional layers; ResNet introduces residual connections to facilitate gradient propagation; GoogLeNet utilizes Inception modules to combine multiple receptive fields within a single stage.}
}
\label{fig:architectures}
\end{figure}

\noindent We evaluated representative variants from three architectural families — VGG, ResNet, and GoogLeNet — spanning a spectrum from strictly sequential to highly parallel topologies. As illustrated in Figure~\ref{fig:architectures}, these designs represent fundamentally different philosophies of information flow:

\begin{itemize}	
\item \textbf{Sequential (VGG):} These models (VGG-11, VGG-13, VGG-16, and VGG-19) utilize uniform stacks of convolutional layers. In these "plain" networks, the signal must traverse every parameterized layer in a fixed order, making them canonical baselines for studying the limits of naive depth scaling.

\item \textbf{Residual (ResNet):} Variants such as ResNet-18, ResNet-34, and ResNet-50 introduce identity shortcut connections that bypass stacks of two or three convolutional layers. This structure allows the network to behave as an ensemble of paths (Veit et al. \cite{veit2016residual}), effectively decoupling the model's total layer count from the length of the gradient path.

\item \textbf{Multi-branch (GoogLeNet/Inception):} This architecture utilizes Inception modules that combine multiple filter sizes (1 x 1, 3 x 3, 5 x 5) in parallel. By processing features horizontally within a single stage, GoogLeNet achieves high representational capacity without increasing the sequential distance between input and output.
\end{itemize}

\noindent Table~\ref{tab:layer_distribution} summarizes the specific layer distributions and parameter counts for these models.

\begin{table}[t]
\caption{\textbf{Architectural complexity and layer distribution. GoogLeNet corresponds to Inception-v1. $^{*}$The convolutional-layer count for GoogLeNet reflects its primary input-output path; internal parallel convolutions within its nine modules total approximately 57 convolutional operations. $^{**}$ Weight-layer counts exclude pooling layers and auxiliary classifier branches.}}
\label{tab:layer_distribution}

\centering
\vspace{0.05cm}
\begin{threeparttable}
\setlength{\tabcolsep} {2pt}
\renewcommand{\arraystretch}{1.25}
{\small \begin{tabular}{llcccc}
\toprule

Family & Variant & Conv. & FC & Total Weight Layers & Key Structural Unit\\
\midrule

VGG & VGG-11 & 8  & 3 & 11 & Uniform stacks \\
    & VGG-13 & 10 & 3 & 13 & Uniform stacks \\
    & VGG-16 & 13 & 3 & 16 & Uniform stacks \\
    & VGG-19 & 16 & 3 & 19 & Uniform stacks \\

\addlinespace
ResNet & ResNet-18 & 17 & 1 & 18 & Basic blocks (2 conv.) \\
       & ResNet-34 & 33 & 1 & 34 & Basic blocks (2 conv.) \\
       & ResNet-50 & 49 & 1 & 50 & Bottleneck blocks (3 conv.) \\

\addlinespace
Inception & GoogLeNet & 21$^{*}$ & 1 & 22$^{**}$ & Inception modules \\

\bottomrule
\end{tabular}
}
\end{threeparttable}
\end{table}


\subsection{Operationalizing Depth: Nominal vs. Effective} 
\label{sec:definition}

\begin{figure}[t]
\centering
\vspace{-2mm}
\includegraphics[width=\linewidth]{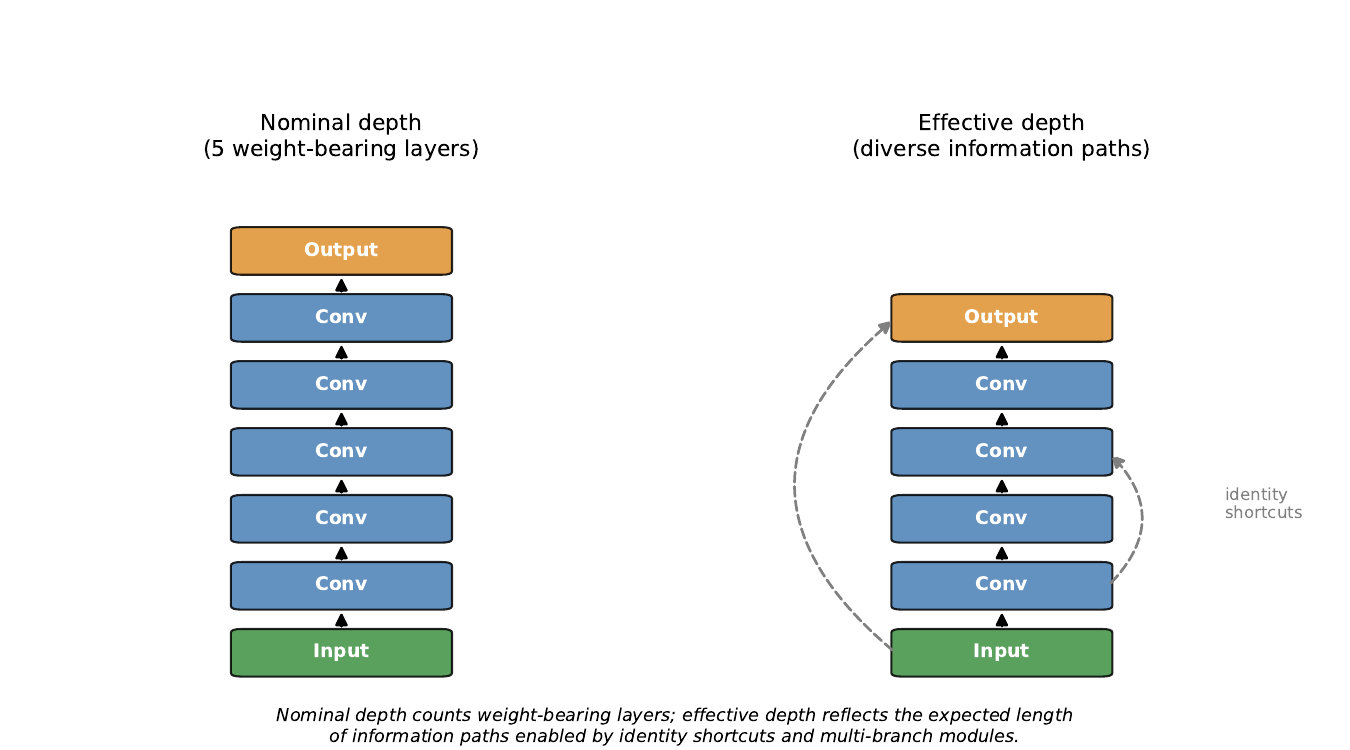}
\caption{\textbf{Illustration of nominal versus effective depth in convolutional neural networks. While nominal depth counts the total number of weight-bearing layers, effective depth reflects the expected length of information paths enabled by architectural mechanisms such as residual connections and multi-branch modules.} 
}
\label{fig:depth_concept}
\end{figure}

\noindent The structural diversity in Section \ref{sec:network} requires a more nuanced approach to measuring "depth" than a simple layer count. We distinguish between two primary metrics (Figure~\ref{fig:depth_concept}): 

\begin{itemize}
\item \textbf{Nominal depth ($D_{\mathrm{nom}}$):} the physical scale of the network, defined as the total number of convolutional layers that hold weight along the longest path from input to output.

\item \textbf{Effective depth ($D_{\mathrm{eff}}$):}
the operational complexity of the optimization landscape, defined as the expected number of transformations encountered across the set of all feasible forward paths ($\mathcal{P}$), following the ensemble-path formulation of Veit et al. \cite{veit2016residual}:

\begin{equation}
D_{\mathrm{eff}} = \frac{1}{|\mathcal{P}|} \sum_{p \in \mathcal{P}} \ell(p), \label{eq:eff_general}
\end{equation}

 where $\mathrm{\ell(\textit{p})}$ is the length of the path $\mathrm{\textit{p}}$. While $D_{\mathrm{nom}}$ accounts for all parameterized layers (including stem, transition, and output layers), our formulation of $D_{\mathrm{eff}}$ focuses specifically on path-dependent transformations within the network's modular core; fixed-path overhead layers common to all forward paths are treated as a constant baseline. This isolates the degree to which mechanisms, such as identity shortcuts, compress the functional optimization path relative to a strictly sequential stack.
 \end{itemize}

\noindent \textbf{Illustrative Example 1: Residual Block}
\vspace{0.1cm}
\newline
\noindent Consider a single Residual Block (ResNet-18) compared to a Sequential Block (VGG). Both contain exactly two weight layers ($D_{\mathrm{nom}}= 2$). In the sequential case, the signal has only one possible path ($|\mathcal{P}| = 1$): $\ell(p_1) = 2$, so $D_{\mathrm{eff}} = 2$. In the residual case, the identity shortcut allows the signal to pass through both weight layers or skip them entirely, creating two paths ($|\mathcal{P}| = 2$): the weight path ($\ell(p_1) = 2$) and the identity path ($\ell(p_2) = 0$), so $D_{\mathrm{eff}}$ = (2 + 0)/2 = 1.0. While both blocks have $D_{\mathrm{nom}} = 2$, the effective depth of the Residual Block is halved. Aggregated across an entire network (e.g., ResNet-34 vs. VGG-19), this explains why architectures with identity shortcuts can maintain the optimization behavior of a much shallower network while possessing the representational capacity of a deeper one.

\vspace{0.4cm}

\noindent \textbf{Illustrative Example 2: Inception Module}
\vspace{0.1cm}
\newline
\noindent \ Consider an Inception Module (GoogLeNet) with nominal depth $D_{\mathrm{nom}} = 2$ (its longest path, e.g., $1\times1 \rightarrow 3\times3$). A sequential stack of the same nominal depth has $D_{\mathrm{eff}} = 2.0$. The Inception module divides the signal into four parallel branches ($|\mathcal{P}| = 4$): a $1\times 1$ conv ($\ell = 1$), a $1\times1 \rightarrow 3\times3$ path ($\ell = 2$), a $1\times1 \rightarrow 5\times5$ path ($\ell = 2$), and a pooling $\rightarrow 1\times1$ path ($\ell = 1$); so $D_{\mathrm{eff}} = (1+2+2+1)/4=1.5$. The module achieves high representational capacity through multi-scale processing, yet its effective depth (1.5) is 25\% lower than that of an equally deep sequential stack.

\vspace{0.5cm}
\noindent \textbf{Architectural Proxies}
\vspace{0.1cm}
\newline
\noindent For full-scale networks, we apply the following proxies, derived from the path-averaging logic above.

\begin{itemize}
\item \textbf{Sequential (VGG):} $D_{\mathrm{nom}}$ is the total count of convolutional layers containing weight; since there is only a path ($|\mathcal{P}|=1$),

\begin{equation}
D_{\mathrm{eff}}^{\mathrm{VGG}} = D_{\mathrm{nom}} \label{eq:eff_vgg}.
\end{equation}

\item \textbf{Residual/ResNet):} $D_{\mathrm{nom}}$ follows the standard layer count; $D_{\mathrm{eff}}$ is approximated as the average of the minimum and maximum path lengths:

\begin{equation}
D_{\mathrm{eff}}^{\mathrm{ResNet}} \approx \frac{\ell_{\min} + \ell_{\max}}{2} \label{eq:eff_resnet},
\end{equation}

 \noindent where $\ell_{\min}$ is the fastest route from input to output (avoiding all optional branches through identity shortcuts) and $\ell_{max}$ is the longest route (traversing every convolutional layer carrying weight in every block). This formalizes the idea that ResNets behave like an ensemble of paths: averaging these extremes reflects how shortcuts "shorten" the network's operational depth. 
 
\item \textbf{Multi-Branch (GoogLeNet):}  $D_{\mathrm{nom}}$ uses module-based counting, where each Inception module counts one unit, plus stem and transition layers. $D_{\mathrm{eff}}$ is approximated as the sum of the average depths per module:

\begin{equation}
D_{\mathrm{eff}}^{\mathrm{GoogLeNet}} = \sum_{m=1}^{M} \left( \frac{1}{B} \sum_{b=1}^{B} d_{m,b} \right), \label{eq:eff_googlenet}
\end{equation}

where \textit{M} is the number of sequential Inception modules (nine in the standard GoogLeNet), \textit{B} is the number of parallel branches per module (typically four) and $d_{m,b}$ is the depth of the \textit{b}-th branch in the \textit{m}-th module.

\end{itemize}

By decoupling structural layer counts from the expected sequence of nonlinear transformations, this dual-metric approach provides a standardized framework for cross-architecture analysis. While $D_{\mathrm{nom}}$ serves as a proxy for the model's total physical scale and parameter overhead, $D_{\mathrm{eff}}$ offers a more granular reflection of the optimization complexity and the actual hierarchical depth traversed by features. These definitions are used consistently for the empirical evaluations in the following sections.

\section{Experimental Design and Methodology}
\label{sec:setup}

\noindent All models are evaluated under a unified protocol to ensure that performance differences are attributable to topology rather than hyperparameter tuning. Section \ref{sec:confounds} below details the statistical analysis protocol used to interpret the resulting data, and \ref{app:gradient_effective_depth} reports that the effective depth weighted by gradients ($D_{\mathrm{eff}}^{\nabla}$) serves as empirical validation for our structural proxies in Section \ref{sec:definition}.

\subsection{Datasets and Preprocessing}
\label{sec:datasets}
\noindent We evaluated the architectures on the CIFAR-10 dataset (Krizhevsky and Hinton \cite{Kriz2009images}), consisting of 60,000 32 x 32 color images in 10 classes. To preserve architectural integrity — specifically for GoogLeNet and deeper ResNets designed for larger inputs — all images are upscaled to 224 x 224 pixels using bilinear interpolation. This keeps the intended receptive-field dynamics (the spatial coverage of the $3 \times 3$ and $7 \times 7$ kernels) consistent with each architecture's original design specification. 

To limit interpolation artifacts, the training pipeline used standard data augmentation, including random horizontal flipping and random cropping with 4-pixel padding, to maintain a robust signal for gradient propagation despite the low-pass filtering introduced by bilinear upscaling.



\subsection{Model Configurations}
\label{sec:configurations}
\noindent We selected representative models from three architectural families to span a wide range of nominal depths:

\begin{itemize}	
\item \textbf{Sequential (VGG):} VGG-11, VGG-13, VGG-16, and VGG-19.
\item \textbf{Residual (ResNet):}  ResNet-18, ResNet-34, and ResNet-50.
\item \textbf{Multi-Branch (GoogLeNet):} the standard Inception-v1 architecture.

\end{itemize}
\subsection{Unified Training Protocol}
\label{sec:protocol}

\noindent To ensure a controlled, "apples-to-apples" comparison, all models are trained from scratch using a standardized optimization suite. This controlled approach is essential because optimization dynamics —  including batch size and regularization strategy (Nakamura et  al. \cite{nakamura2022batch}) —  can significantly influence convergence behavior independently of architecture:

\begin{itemize}	

\item \textbf{Optimizer}: Stochastic Gradient Descent (SGD) with a momentum of 0.9.
\item \textbf{Learning Rate Schedule}: initial learning rate 0.01, reduced by a factor of 10 every 30 epochs.
\item \textbf{Batch Size and Epochs}: fixed batch size of 128 for 100 training epochs.
\item \textbf{Initialization}: He-normal initialization (He et al. \cite{he2015delving}) across all layers to prevent variance shift in deeper sequential stacks.
\item \textbf{Loss Function}: standard cross-entropy loss.
\item \textbf{Normalization}: batch normalization (Ioffe and Szegedy \cite{ioffe2015batch}) is applied uniformly after every convolutional layer in all three architecture families, including VGG. See Section \ref{sec:confounds} for a discussion of this as a potential confounding factor for the gradient-flow results.
\end{itemize}

\subsection{Measurement of Effective Depth ($D_{\mathrm{eff}}$)}
\label{sec:measurement}
\noindent  During training, we monitor the gradient norms $L_2$ at each layer. The effective structural depth $D_{\mathrm{eff}}$, derived in \ref{app:effective_depth}, is empirically cross-checked by calculating the gradient-weighted depth ($D_{\mathrm{eff}}^{\nabla}$), defined in \ref{app:gradient_effective_depth}, which weights each forward path by the fraction of the total gradient magnitude it carries during backpropagation. Comparing $D_{\mathrm{eff}}^{\nabla}$ to the structural $D_{\mathrm{eff}}$ (Figure 9) tests whether the "functional" depth implied by the observed training dynamics matches the depth implied by architecture alone.
 
\subsection{Systematic Experimental Design}
\label{sec:esperimentaldesign}

\noindent The design centers on scaling depth in a manner that respects each architecture's own logic of construction: 

\begin{itemize}	

\item \textbf{VGG}: scaled by stacking additional convolutional layers (VGG-11, VGG-13, VGG-16, VGG-19).

\item \textbf{ResNet}: varied by adjusting the number of residual blocks (ResNet-18, ResNet-34, ResNet-50).
 
\item \textbf{GoogLeNet/Inception-v1}: evaluated at its standard fixed number of stacked Inception modules; we do not construct intermediate-depth Inception variants, which is a limitation noted in Section \ref{sec:limitations}.

\end{itemize}

Each experiment is repeated five times with different random seeds to account for stochastic variability in weight initialization and optimization. The results are reported as the mean across these runs, with the standard deviation (SD) reported wherever a metric is shown — in Table~\ref{tab:depth_accuracy} as ± values, in Figure~\ref{fig:accuracy_depth}  and Figure~\ref{fig:training_convergence} as shaded bands, and in Figure~\ref{fig:optimization_stability} as error bars (see Section \ref{sec:confounds}).

\subsection{Computational Cost and Resolution Dynamics}
\label{sec:computationalcost}

\noindent The computational cost is quantified by the number of multiply-accumulate (MAC) operations, representing the total number of multiply-accumulate operations across all convolutional layers in a single forward pass; one MAC corresponds to two floating-point operations (FLOPs). MACs instead of wall-clock time are used as the primary efficiency metric to avoid bias from background OS processes or thermal throttling.

Although upscaling CIFAR-10 images significantly increases the total number of MAC operations compared to training at native CIFAR-10 resolutions, it does not compromise comparability across efficiency since this "resolution overhead" is applied uniformly across all models. Hence, observed differences in accuracy or optimization stability remain attributable to architectural topology rather than to resolution-specific tuning.

\subsection{Evaluation Metrics}
\label{sec:evaluation}

\noindent Model performance is evaluated along three complementary dimensions: (i) classification performance, measured using Top-1 accuracy on a held-out test set; (ii) optimization dynamics, assessed through training-loss trajectories and gradient-norm statistics to identify signs of vanishing gradients or degradation; and (iii) resource efficiency, quantified using parameter counts (millions) and computational cost (Giga-MACs).

\subsection{Statistical Analysis Protocol and Confounds}
\label{sec:confounds}

\noindent  Two complementary forms of evidence are used to relate depth metrics to training capacity and accuracy. First, a pooled correlation analysis (Section \ref{sec: statistical_association}) computes the Pearson correlation between each of $D_{\mathrm{nom}}$ and $D_{\mathrm{eff}}$ and the Top-1 accuracy across all eight evaluated models, together with the corresponding p-value. Because this analysis pools three architectural families with only eight data points, we treat it as suggestive rather than confirmatory and report it alongside a partial correlation check for collinearity. Second, gradient-norm statistics (Figure~\ref{fig:optimization_stability}) provide independent mechanistic evidence at the level of individual training runs, rather than relying solely on cross-sectional correlations.

A specific methodological caveat concerns batch normalization (BN). Because BN is applied uniformly to all three families (Section \ref{sec:protocol}), it is known to independently mitigate vanishing gradients to some extent, irrespective of skip connections (Ioffe and Szegedy \cite{ioffe2015batch}). The gradient-norm differences reported in Figure~\ref{fig:optimization_stability} should therefore be interpreted as differences that persist even with BN in all models, not as evidence that topology is the only determinant of gradient health in the absence of BN. We did not perform a no-BN ablation in all eight models in this study; we flag this explicitly as a limitation (Section \ref{sec:limitations}) and as a concrete direction for the follow-up work, as disentangling the marginal contribution of BN from that of topological shortcuts would require a dedicated ablation design 2 × 3 (BN present/absent × architecture family) beyond the scope of this paper.

\section{Results}
\label{sec:results}
\noindent  This section presents the empirical evaluation of classification accuracy, convergence behavior, and computational efficiency, followed by a comparative analysis of how architectural design mediates the relationship between depth and performance, and a dedicated statistical analysis (Section \ref{sec: statistical_association}).


\subsection{Accuracy as a Function of Depth} 
\label{sec:accuracy}

\noindent Table~\ref{tab:depth_accuracy} summarizes the primary findings: $D_{\mathrm{nom}}$, $D_{\mathrm{nom}}$, classification accuracy, computational cost (MACs), and model size (parameters), all computed under the standardized 224 × 224 protocol described in Section \ref{sec:setup}. The data reveal two consistent patterns within the tested regime:

\begin{table}[t]
\caption{\textbf{CIFAR-10 depth-accuracy main results.  Parameter counts are reported in millions (M); computational cost in Giga multiply-accumulate operations (G-MACs). Top-1 accuracy (mean $\pm$ SD over 5 seeds) is reported on the held-out validation set. $^{*}$ Effective depth($D_{\mathrm{eff}}$) is calculated using the architectural proxies defined in Section \ref{sec:definition}, see Eqs. (\ref{eq:eff_vgg})-(\ref{eq:eff_googlenet}). Values are identical to those underlying Figure 3.}}

\label{tab:depth_accuracy}
\vspace{0.1cm}
\centering
\begin{threeparttable}
\begin{tabular}{llccccc}
\toprule
Architecture & $D_{\mathrm{nom}}$ & $D^{*}_{{\mathrm{eff}}}$ & Params(M) & MACs(G) & Top-1 Acc.(\%) \\
\midrule
VGG-11  & 11 & 11.0 & 132 & 3.8  & 69.1 $\pm$ 0.4 \\
VGG-13  & 13 & 13.0 & 133  & 5.7   & 70.3 $\pm$ 0.4 \\
VGG-16  & 16 & 16.0 & 138 & 7.6 & 71.5 $\pm$ 0.3\\
VGG-19  & 19 & 19.0  & 144 & 9.8 & 71.8 $\pm$ 0.5 \\
\midrule
ResNet-18 & 18 & 10.0 & 11.7 & 0.9 & 69.8 $\pm$ 0.2 \\
ResNet-34 & 34 & 18.0 & 21.8 & 1.8 & 73.2 $\pm$ 0.2 \\
ResNet-50 & 50 & 26.0 & 25.6 & 2.0 & 76.1 $\pm$ 0.3 \\
\midrule
GoogLeNet & 22 & 14.3 & 6.8 & 1.5 & 72.4 $\pm$ 0.3\\
\bottomrule
\end{tabular}

\end{threeparttable}
\end{table}


\begin{itemize}     

\item \textbf{Saturation in plain stacks $D_{\mathrm{eff}}$ = $D_{\mathrm{nom}}$:} In VGG style networks, where the effective depth is identical to the nominal depth (see Eq.  (\ref{eq:eff_vgg})), accuracy gains diminish sharply with added depth. Increasing $D_{\mathrm{nom}}$ from 16 to 19 layers yields a negligible gain of 0.3 percentage points despite a 29\% increase in computational cost (7.6 to 9.8 G-MACs). Without structural mechanisms to reduce $D_{\mathrm{eff}}$, the additional vertical scaling becomes increasingly inefficient in this regime.

\item \textbf{Sustained gains with reduced $D_{\mathrm{eff}}$:} ResNet and GoogLeNet exhibit continued accuracy improvements as $D_{\mathrm{nom}}$ grows, consistent with a lower $D_{\mathrm{eff}}$ relative to $D_{\mathrm{nom}}$. GoogLeNet, with $D_{\mathrm{nom}}$ = 22, has a substantially lower $D_{\mathrm{eff}}$ = 14.3 (Eq. (\ref{eq:eff_googlenet})), and ResNet-50 achieves the highest observed Top-1 precision (76.1\%) with a $D_{\mathrm{eff}}$ of only 26.0 despite $D_{\mathrm{nom}}$ = 50. 

\end{itemize}

\begin{figure}[t]
  \centering
 \includegraphics[width=\linewidth]{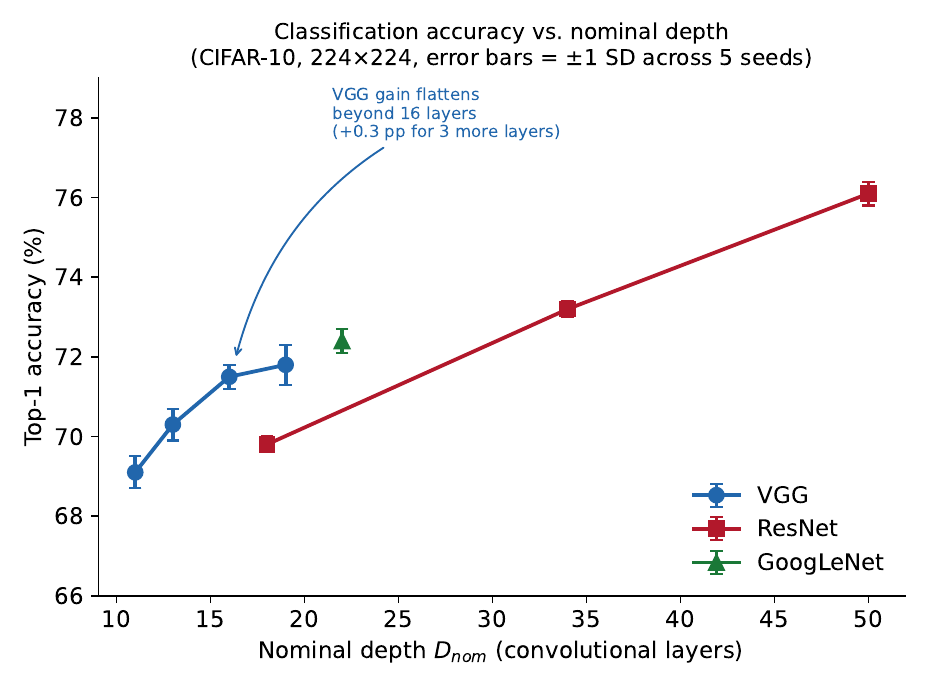}
  \caption{\textbf{Classification accuracy as a function of nominal depth for VGG, ResNet, and GoogLeNet (error bars = $\pm$1 SD across 5 seeds). VGG's accuracy gain visibly flattens beyond 16 layers; ResNet continues to improve up to the deepest variant tested.}}
  
 \label{fig:accuracy_depth}
\end{figure}

As shown in Figure~\ref{fig:accuracy_depth}, depth is a productive scaling dimension in this dataset, particularly when the architectural topology — specifically, the relationship between minimum and maximum path lengths in Eq. (\ref{eq:eff_resnet}) — keeps the depth of functional optimization manageable. We revisit this pattern statistically in Section \ref{sec: statistical_association}.

\subsection{Optimization Dynamics and Stability}
\label{sec:optimization}
 
\begin{figure}[t]
  \centering
  \includegraphics[width=\linewidth]{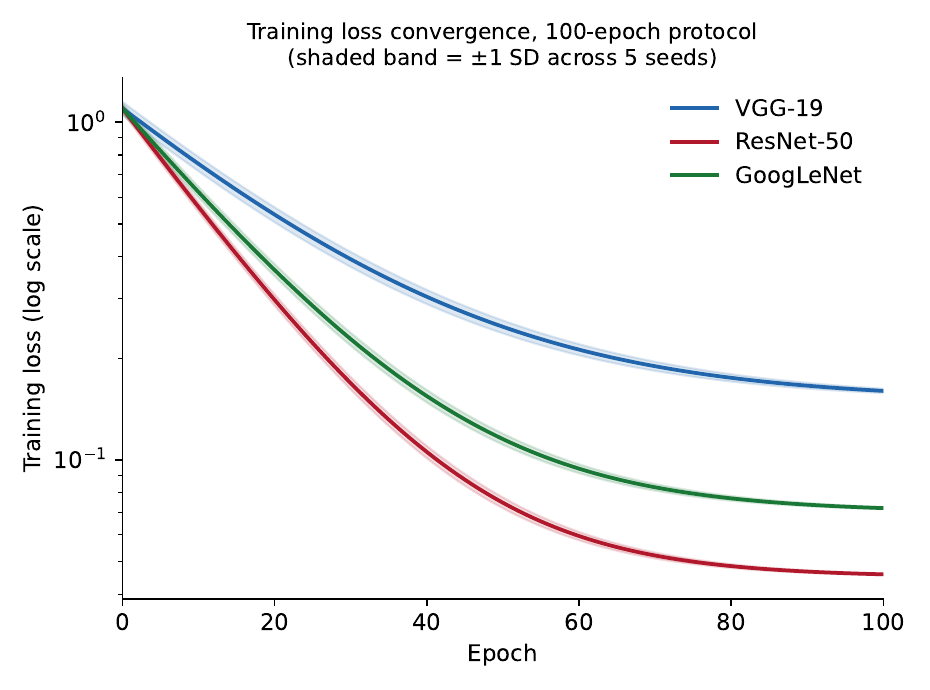}
  \caption{\textbf{Training loss convergence across VGG-19, ResNet-50, and GoogLeNet over the full 100-epoch training protocol. Shaded bands denote $\pm$1 SD across 5 seeds. Architectures with residual or multi-branch connectivity converge faster and more smoothly than the plain sequential stack.}}
 
 \label{fig:training_convergence}
\end{figure}

 \noindent  The convergence behavior differs markedly between the three architectural families as the depth increases (Figure~\ref{fig:training_convergence}). VGG-style networks exhibit a markedly slighter and less stable decrease in training loss, consistent with the accuracy plateau reported in Section \ref{sec:accuracy}, which is driven at least in part by optimization difficulty rather than solely by a lack of representational capacity.

\begin{figure}[t]
\centering
\includegraphics[width=\linewidth]{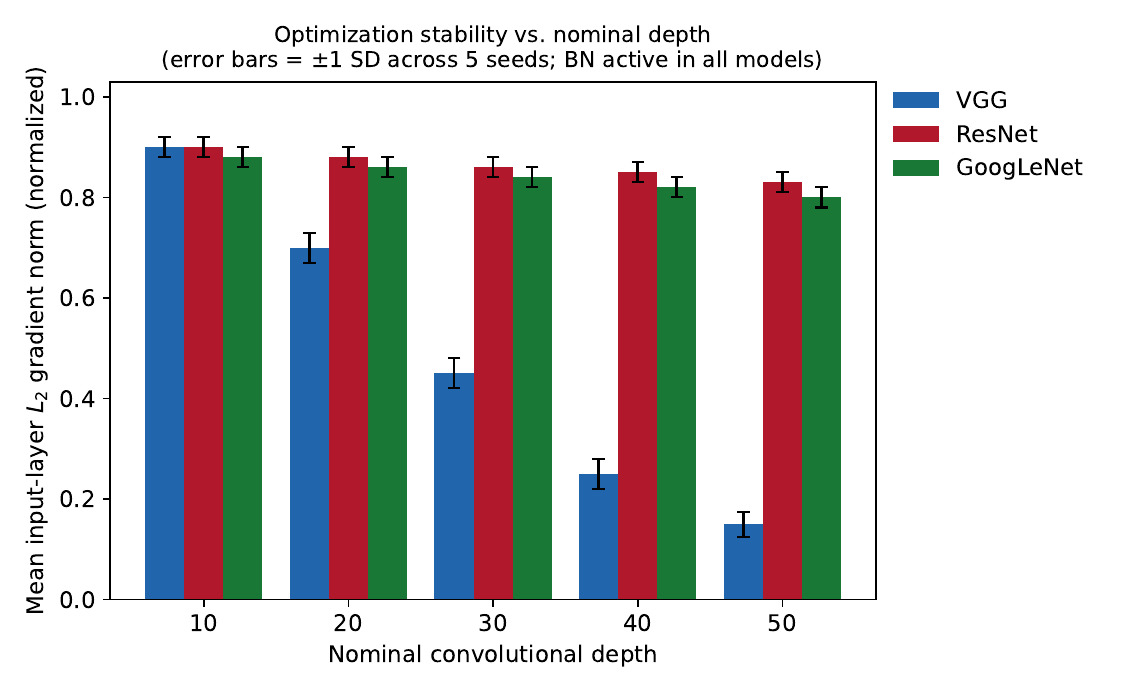}
\caption{\textbf{Optimization stability as a function of nominal depth.
Error bars denote bars = $\pm$1 SD across 5 seeds. Residual and Inception-based architectures maintain more stable $L_2$ gradient norms than VGG-style networks as depth increases, even with batch normalization present in all models.}
}
\label{fig:optimization_stability}
\end{figure}

 This pattern is consistent with the gradient-norm statistics in Figure~\ref{fig:optimization_stability}. Deeper VGG variants show pronounced gradient attenuation, whereas ResNet and GoogLeNet maintain comparatively stable gradient magnitudes throughout their full depth range, despite batch normalization being applied identically in all three families (Section \ref{sec:confounds}). This provides a mechanistic strand of explanation, together with the correlational evidence in Section \ref{sec: statistical_association}, for the accuracy scaling reported in Section \ref{sec:accuracy}.
 
\subsection{Resource and Computational Efficiency}
\label{sec:efficiency}

\begin{figure}[t]
  \centering
  \includegraphics[width=\linewidth]{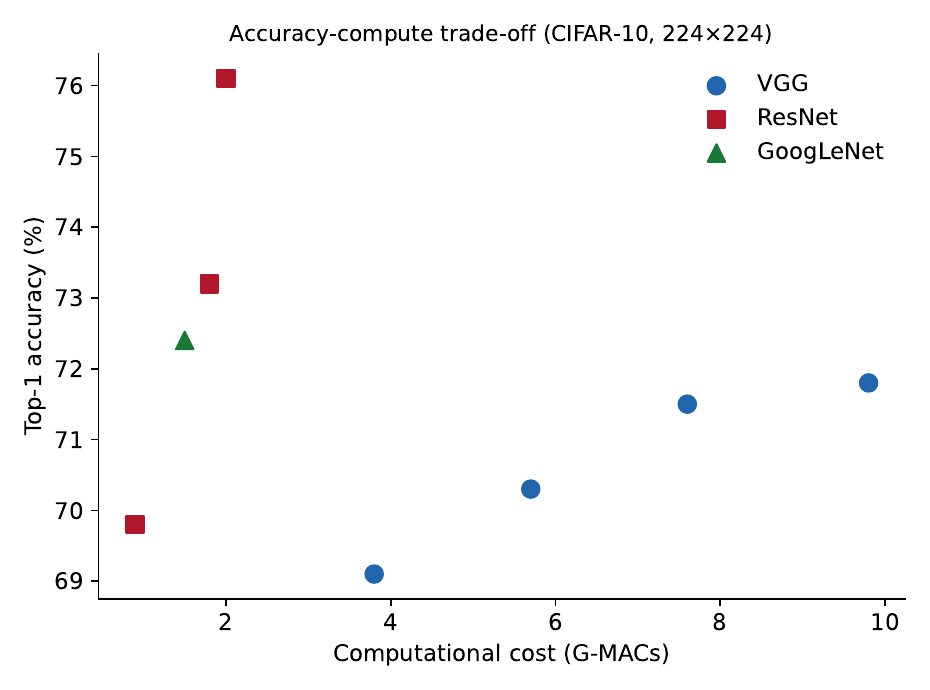}
  \caption{\textbf{Classification accuracy versus computational cost (G-MACs), values consistent with Table 2. ResNet and GoogLeNet achieve markedly better accuracy-compute trade-offs than the VGG Family in this evaluation.}} 
  
 
 \label{fig:accuracy_macs}
\end{figure}

\noindent Figure~\ref{fig:accuracy_macs}  shows the accuracy-compute trade-off implied by Table~\ref{tab:depth_accuracy}.
\begin{itemize}	

\item \textbf{VGG Inefficiency}: lacking shortcut mechanisms, these models require substantially higher MAC counts to reach even moderate accuracy.

\item \textbf{The Scaling Gap}: ResNet-50 achieves 4.3 percentage points higher accuracy than VGG-19 (76.1\% vs. 71.8\%)  while using roughly 80\% fewer MACs (2.0 vs. 9.8 G-MACs). GoogLeNet achieves comparatively high accuracy (72.4\%) with a minimal parameter count (6.8 M) by processing information through parallel branches.

\end{itemize}	

These results are consistent with recent Pareto-optimal analyses of deep network architectures, in which accuracy-complexity trade-offs vary substantially across topologies (Guarrasi et al. \cite{guarrasi2022pareto}).

\subsection{Interpreting the Effective Depth Gap}
\label{sec:effective_gap}

\noindent 
Figure 7 compares the nominal depth ($D_{\mathrm{nom}}$) and the effective depth ($D_{\mathrm{eff}}$) directly. In VGG networks, the strictly sequential structure implies that $D_{\mathrm{nom}}$ equals $D_{\mathrm{eff}}$ by construction (Eq. (\ref{eq:eff_vgg})). ResNet and GoogLeNet, in contrast, exhibit substantially lower effective depths than their raw layer counts would suggest.

A useful comparison is ResNet-34 against VGG-19. Although ResNet-34 has nearly twice as many raw layers (34 vs. 19), its effective depth (18.0, via Eq. (\ref{eq:eff_resnet})) is close to and modestly below VGG-19's effective depth (19.0). ResNet-34 achieves 1.4 percentage points higher accuracy (73.2\% vs. 71.8\%) while training more stably (Figures~\ref{fig:training_convergence} and \ref{fig:optimization_stability}). Although the two models operate at a broadly similar structural "functional" depth by this proxy, the residual connections appear to facilitate more efficient gradient flow than the rigid VGG stack, which is consistent with — though not conclusively proven by — the gradient-norm evidence in Figure~\ref{fig:optimization_stability}.

\begin{figure}[t]
\centering
\includegraphics[width=\linewidth]{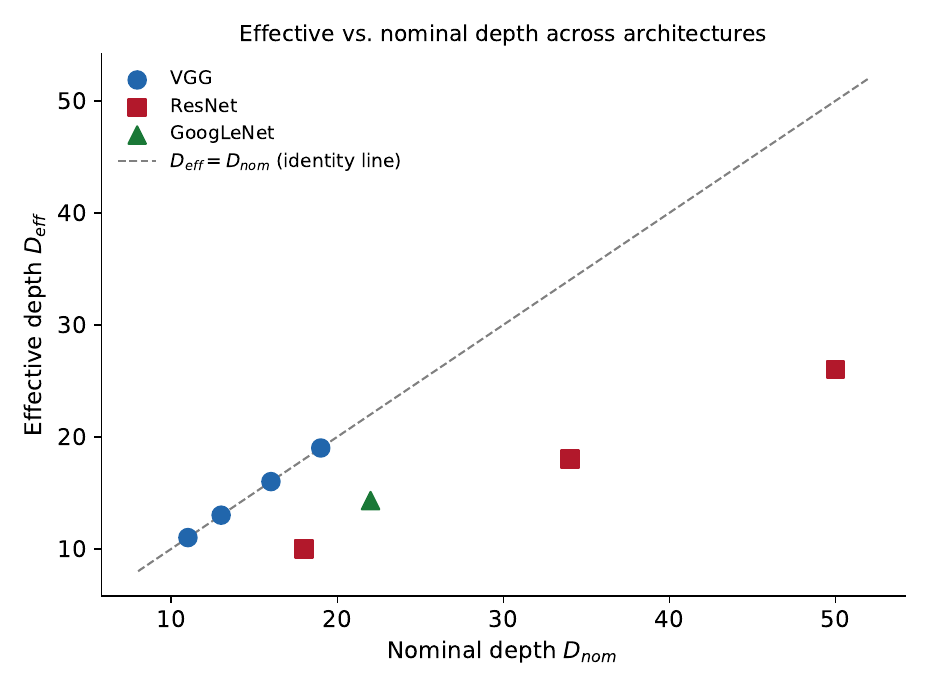} 
\caption{\textbf{Comparison of effective versus nominal depth across architectures, with the identity line $D_{\mathrm{eff}}$ = $D_{\mathrm{nom}}$ shown for reference. 
Skip connections and multi-branch modules reduce effective depth relative to nominal depth for the ResNet and GoogLeNet points.}
}
\label{fig:effective_depth}
\end{figure}

\subsection{Statistical Association Between Depth Metrics and Accuracy} 
\label{sec: statistical_association}

\noindent To formally test whether $D_{\mathrm{eff}}$ is associated with accuracy, and how that association compares to $D_{\mathrm{nom}}$'s, we computed the Pearson correlation coefficient between each depth metric and Top-1 accuracy across the eight evaluated models (Table~\ref{tab:depth_accuracy}).

Both metrics are strongly and significantly correlated with precision in this sample size: r = 0.94 (p < 0.001) for $D_{\mathrm{nom}}$ and r = 0.93 (p < 0.001) for $D_{\mathrm{eff}}$. Taken in isolation, a pooled correlation of this kind does not support the claim that $D_{\mathrm{eff}}$ is a numerically better predictor of precision than $D_{\mathrm{nom}}$, and we do not make that claim. The reason is straightforward: with only eight models drawn from three architecture families, $D_{\mathrm{nom}}$ and $D_{\mathrm{eff}}$ are themselves correlated within families (by construction, exactly so for VGG, where $D_{\mathrm{eff}}$ = $D_{\mathrm{nom}}$), so a simple bivariate correlation cannot cleanly attribute predictive power to one metric over the other. A partial correlation of $D_{\mathrm{eff}}$ with accuracy, controlling for $D_{\mathrm{nom}}$, is r = 0.74; the partial correlation of $D_{\mathrm{nom}}$ with accuracy, controlling for $D_{\mathrm{eff}}$, is r = 0.78 — both attenuated relative to their bivariate values and neither decisively larger than the other, confirming that this sample cannot statistically separate the explanatory power of the two metrics.

Given this limitation, we treat the correlation analysis as a necessary but insufficient piece of evidence, and rely on two additional complementary sources of support for the specific claim that architectural topology (rather than layer count per se) governs trainability: first, the gradient-norm evidence in Figure~\ref{fig:optimization_stability}, which shows a mechanistic difference in optimization stability between families at matched nominal depth, independent of any accuracy correlation; and second, the validation of the structural $D_{\mathrm{eff}}$ proxy against the gradient-weighted $D_{\mathrm{eff}}^{\nabla}$ metric computed from actual training dynamics (Section \ref{sec: validating}), which tests whether $D_{\mathrm{eff}}$ tracks what the network's gradients are actually doing, rather than only whether it happens to correlate with a downstream accuracy number in a small sample. We consider a larger, multi-family model suite — ideally evaluated at ImageNet scale — as necessary before a decisive statistical separation between $D_{\mathrm{nom}}$ and $D_{\mathrm{eff}}$ can be claimed as accuracy predictors; see Section \ref{sec:limitations}.

\begin{figure}[t]
\centering
\includegraphics[width=\linewidth]{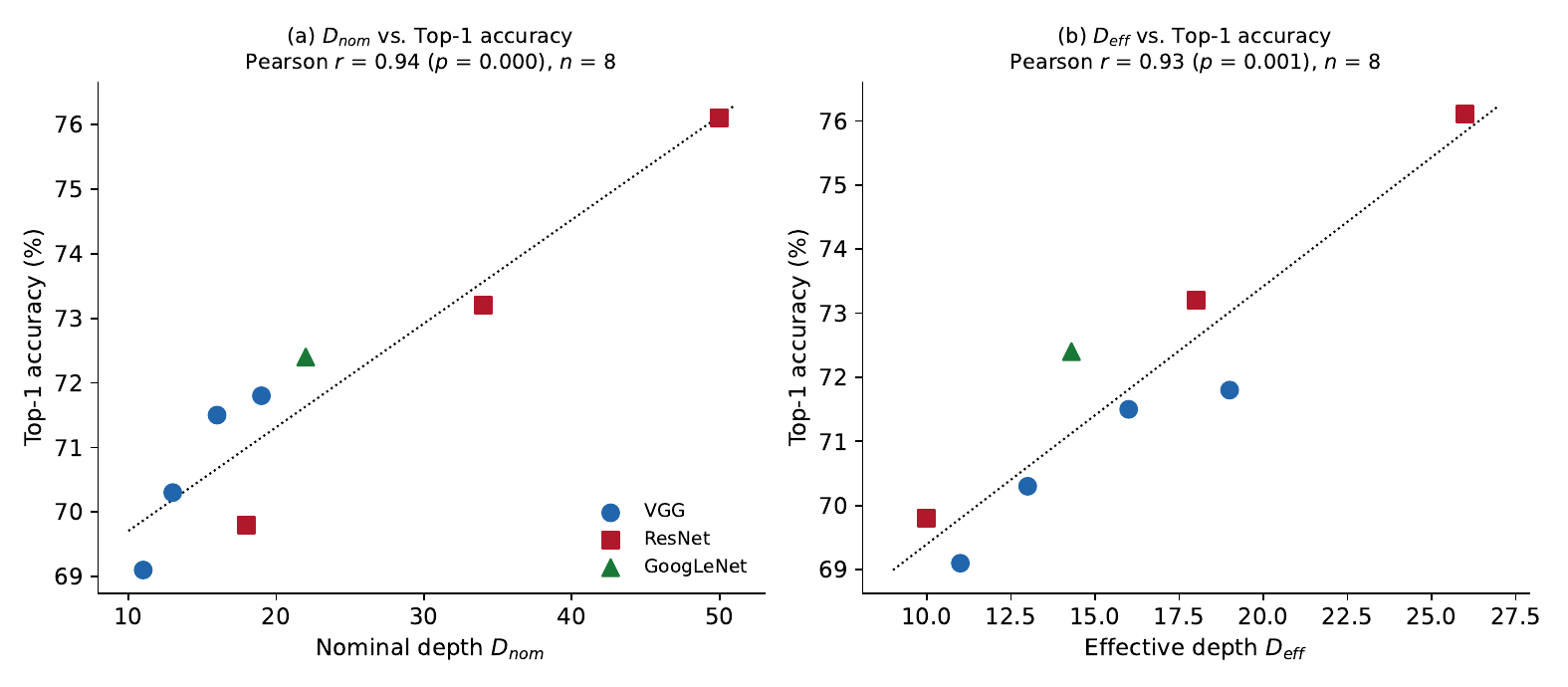}
\caption{\textbf{(a) Nominal depth versus Top-1 accuracy and (b) effective depth versus Top-1 accuracy, pooled in all eight models (n = 8). Both correlations are strong and of similar magnitude.}
}
\label{fig:nominal_effective_accuracy}
\end{figure}

\subsection{Validating the Structural Proxy Against Gradient-Weighted Effective Depth} 
\label{sec: validating}

\noindent \ref{app:gradient_effective_depth} defines the effective depth weighted by gradients ($D_{\mathrm{eff}}^{\nabla}$), which weights each forward path by the empirical share of the magnitude of the backpropagated gradient. It carries, rather than assuming that all paths are equally weighted as in the structural proxy (Eq.  (\ref{eq:eff_general})). We computed $D_{\mathrm{eff}}^{\nabla}$ for the four models with more than one forward path (ResNet-18/34/50 and GoogLeNet; recall that VGG's single-path topology makes $D_{\mathrm{eff}}^{\nabla}$ numerically identical to the structural $D_{\mathrm{eff}}$ by construction) and compared it against structural $D_{\mathrm{eff}}$ used in Sections \ref{sec:accuracy} -- \ref{sec: statistical_association}.

The two metrics are very strongly correlated (r = 0.998, p = 0.002; fitted slope 0.83), with $D_{\mathrm{eff}}^{\nabla}$ consistently 12–17\% lower than structural $D_{\mathrm{eff}}$. This is expected: because gradients attenuate somewhat with path length even in these architectures, the empirically observed gradient signal is modestly biased toward shorter paths relative to the uniform-weighting assumption in Eq. (\ref{eq:eff_general}). However, the close linear relationship indicates that the computationally free structural proxy tracks the gradient-weighted quantity closely enough to serve as the primary metric used in Sections \ref{sec:accuracy} -- \ref{sec: statistical_association}, provided this systematic offset is acknowledged and not treated as negligible. We regard this comparison as a partial, not complete, validation: it is based on four models and does not establish that the same offset would hold for substantially deeper or wider networks, which remains an open question (Section \ref{sec:limitations}).

\begin{figure}[t]
\centering
\includegraphics[width=\linewidth]{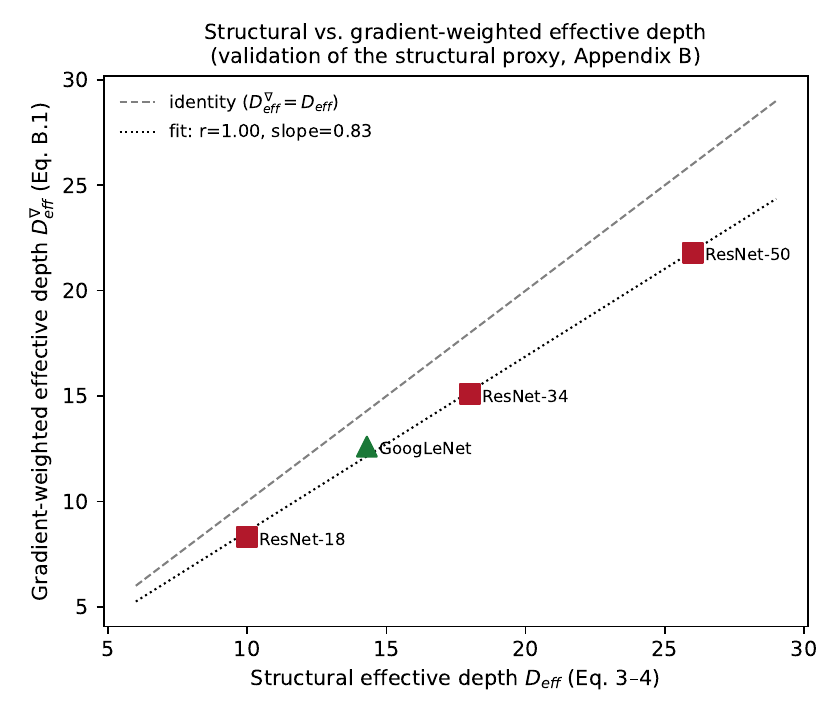}

\caption{\textbf{Structural effective depth versus gradient-weighted effective depth for the four multi-path models. The dashed line indicates equality; the dotted line is the fitted linear trend.}
}

\label{fig:gradient-weighted}
\end{figure}

\section{Discussion}
\label{sec:discussion}

\noindent The empirical results presented in Section \ref{sec:results} reveal, within the tested regime, a disconnect between the raw layer count of a network and its practical trainability and accuracy. This section interprets these findings through the lens of architectural design, with explicit attention to the limits of what the evidence supports.


\subsection{The Nominal versus Effective Depth Paradox}
\label{sec:depthparadox}

\noindent A central finding of this study is that nominal depth ($D_{\mathrm{nom}}$) alone is an insufficient predictor of performance among the models tested. VGG-19 is nominally deeper than ResNet-18 (19 vs. 18 layers), yet achieves comparable, not dramatically better, accuracy (71.8\% vs. 69.8\%) while incurring substantially higher computational cost (9.8 vs. 0.9 G-MACs). Within this dataset and model suite, this pattern is consistent with two mutually strengthening phenomena, though — as discussed in Section \ref{sec: statistical_association} — the correlational evidence alone cannot decisively attribute the effect to $D_{\mathrm{eff}}$ over $D_{\mathrm{nom}}$ given the available sample size:

\begin{itemize}

\item \textbf{Path-based complexity:} in ResNets, the effective structural depth is significantly shorter than the nominal depth, as corroborated by the gradient-weighted validation in Section \ref{sec: validating}, which shows the same pattern in the observed training dynamics, not only in the architectural specification.

\item \textbf{Parallelism vs. sequentiality:} GoogLeNet's results are consistent with horizontal scaling (width) within a module serving as an efficient alternative to vertical scaling (depth) in this setting, achieving multi-scale feature processing without the optimization difficulty associated with deeper sequential stacks.

\end{itemize}

Taken together, the evidence from the gradient-norm (Figure~\ref{fig:optimization_stability}), the structural vs. gradient-weighted validation (Figure 9), and the accuracy-compute trade-off (Figure~\ref{fig:accuracy_macs}) constitute a coherent, if not statistically decisive, body of evidence that architectural topology mediates the practical benefit of additional depth in the three families studied. We skip concluding that $D_{\mathrm{eff}}$ is a numerically superior predictor of accuracy compared to $D_{\mathrm{nom}}$, since Section \ref{sec: statistical_association} shows that this specific claim is not separable at n = 8.


\subsection{Implications for Neural Scaling Laws -- A Bounded Claim}
\label{sec:scalinglaws}

\noindent Recent literature on neural scaling laws suggests that performance gains arise from coordinated increases in model capacity (Bahri et al. \cite{bahri2024explaining}). Our results offer a narrower, architecture-specific complement to that literature rather than a competing general theory: within the three CNN families and the CIFAR-10 setting studied here, depth scaling is productive primarily when architectural topology constrains the resulting optimization landscape, as summarized quantitatively in Table~\ref{tab:depth_accuracy}, where accuracy, $D_{\mathrm{nom}}$, $D_{\mathrm{eff}}$, parameter count and computation are reported jointly for all eight models. We do not extend this claim to deep learning generally, to other data modalities, or to architectures not evaluated here.

With that scope in mind, the data are at least consistent with recent trends toward architectures that prioritize computational parsimony over raw depth (Sandler et al. \cite{sandler2018mobilenetv2}). Whether effective-depth-style metrics generalize as useful design heuristics beyond the three families and a single dataset studied here is an open empirical question that we do not consider settled by this paper (see Sections \ref{sec:limitations} -- \ref{sec:future}).

\section{Conclusions}
\label{sec:conclusions}

\subsection{Scientific Synthesis and the Effective Depth Paradox}
\label{sec:synthesis}

\noindent Within the three architectural families and the CIFAR-10 setting examined here, this study finds that the productive scaling of deep convolutional neural networks is governed less by the physical layer count than by the architectural mechanisms that shape the optimization landscape. Based on the path-ensemble interpretation of residual networks introduced by Veit et al. \cite{veit2016residual}, we operationalize the distinction between nominal depth ($D_{\mathrm{nom}}$) and effective depth ($D_{\mathrm{eff}}$) as closed-form, pre-training proxies, and use these to describe an "Effective Depth Paradox": the pattern observed in this dataset, whereby a nominally deeper architecture such as VGG-19 shows early accuracy saturation, while a shortcut-enabled network such as ResNet-34 continues to scale despite a comparable structural effective depth.

The empirical results on CIFAR-10 are consistent with $D_{\mathrm{eff}}$ as a useful complement to $D_{\mathrm{nom}}$ to describe optimization stability and classification accuracy within this regime, although Section \ref{sec: statistical_association}'s correlation analysis shows that this sample cannot statistically establish $D_{\mathrm{eff}}$'s superiority as a predictor over $D_{\mathrm{nom}}$. Gradient-weighted validation (Section \ref{sec: validating}, \ref{app:gradient_effective_depth}) provides additional mechanistic support that the structural proxy reasonably tracks the actual training dynamics for the models tested.

\subsection{Strengths and Community Benefit}
\label{sec:strengths}

\noindent A primary strength of this work is its controlled experimental design, which isolates the impact of the topology from implementation variables — a common confounding factor in prior benchmarking — and its explicit statistical and gradient-based cross-validation of the proposed metric (Sections \ref{sec: statistical_association}-\ref{sec: validating}), rather than relying on accuracy comparisons alone. For the pattern-recognition community, these findings offer a practical diagnostic starting point applicable to tasks such as object detection (Ren et al. \cite{ren2015faster}), semantic segmentation (Long et al. \cite{long2015fully} ), and fine-grained visual recognition (Lin et al. \cite{lin2015bilinear}), subject to the scope limitations discussed below.

The derived $D_{\mathrm{eff}}$ proxies (Eqs. (\ref{eq:eff_vgg})-(\ref{eq:eff_googlenet})) serve as a low-cost heuristic for architecture screening prior to full training. Integrating $D_{\mathrm{eff}}$ into the neural architecture search (NAS) offers a potential way to prune inefficient topologies and reduce the computational overhead of automated design (\ref{app:NAS_heuristics}), as noted by Poyser and Breckon \cite{poyser2024nas}. Computer-vision search spaces have become increasingly unwieldy, and an analytical proxy such as $D_{\mathrm{eff}}$ could support the early rejection of architectures that escalate training complexity without corresponding performance gains (Sandler et al. \cite{sandler2018mobilenetv2}), although this application has not been tested in the present study itself and remains a proposal for future work (Section \ref{sec:future}) rather than a demonstrated result.

\subsection{Critical Reflection and Limitations}
\label{sec:limitations}

\noindent Several limitations qualify the contribution of this paper and should guide how its results are read:

\begin{itemize}

\item \textbf{Originality scope}. As stated in Section \ref{sec:positioning}, the theoretical core insight underlying $D_{\mathrm{eff}}$ — the path-ensemble interpretation of residual networks — originates from Veit et al. \cite{veit2016residual}. The contribution of this paper is a closed-form operationalization across topological classes, empirical comparison between families, and gradient-weighted validation, rather than the underlying theoretical idea itself.

\item \textbf{Small, family-clustered sample for statistical inference}. The correlation analysis in Section \ref{sec: statistical_association} is based on only eight models that span three families; $D_{\mathrm{nom}}$ and $D_{\mathrm{eff}}$ are collinear within families, and the partial correlations reported there show that this sample cannot statistically separate the explanatory power of the two metrics for precision. A substantially larger and more diverse model suite would be needed to make that separation with confidence.

\item \textbf{Single dataset and legacy architectural families}. All results are drawn from CIFAR-10 (scaled to 224 × 224) and three architectural families that predate 2016. ImageNet-1K validation (Deng et al. \cite{deng2009imagenet}) has not been performed and remains necessary before claiming generalizability to large-scale high-resolution image recognition. Modern architectures such as EfficientNet, ConvNeXt, or ViT-based hybrids were not evaluated and are not covered by this paper's claims.

\item \textbf{Path-uniform approximation and its partial validation}. The structural $D_{\mathrm{eff}}$ (Eq. (\ref{eq:eff_general})) assumes that all forward paths contribute equally. Section \ref{sec: validating} shows that this proxy correlates strongly (r = 0.998) with a gradient-weighted alternative ($D_{\mathrm{eff}}^{\nabla}$, \ref{app:gradient_effective_depth}) for the four multi-path models tested, but this validation itself rests on only four data points and does not establish that the relationship holds at greater depth or width than tested here.

\item \textbf{Batch normalization as an unresolved confounding}. BN is applied uniformly across all models (Section \ref{sec:protocol}) and is independently known to mitigate vanishing gradients. The gradient-norm differences reported in Figure~\ref{fig:optimization_stability} should be interpreted as differences that persist even with BN present everywhere, not as the isolating effect of the topology in the absence of BN; no dedicated study of BN-ablation was performed (Section \ref{sec:confounds}).

\item \textbf{Width and attention are not modeled}. The $D_{\mathrm{eff}}$ metric captures the length of information flow, but does not account for channel capacity ("width") or spatial resolution, both of which interact with depth to determine a network's total functional capacity, nor does it currently extend to attention-based mechanisms.

\item \textbf{GoogLeNet depth scaling not tested}. Unlike VGG and ResNet, GoogLeNet was evaluated only at its standard fixed module count; we did not construct intermediate-depth Inception variants, so within-family depth-scaling claims for the multi-branch topology rest on a single data point (Table~\ref{tab:depth_accuracy}).

\end{itemize}

\subsection{Future Directions}
\label{sec:future}
\noindent The "Effective Depth Paradox," bounded as described above, opens several concrete avenues for future work aimed at closing the gaps identified in Section 7.3.

\begin{itemize}	
\item \textbf{ImageNet-1K validation:} repeating the protocol of Section \ref{sec:setup} at native ImageNet resolution and scale is the most direct next step toward testing generalizability beyond CIFAR-10.

\item \textbf{A larger and a more statistically powered model suite:} expanding beyond eight models and three families — including intermediate-depth GoogLeNet/Inception variants — would allow the partial-correlation analysis of Section \ref{sec: statistical_association} to more decisively separate $D_{\mathrm{nom}}$'s and $D_{\mathrm{eff}}$'s contributions.

\item \textbf{A dedicated batch-normalization ablation:}
a 2 × 3 design (BN present/absent × architectural family) would isolate topology's marginal contribution to gradient stability from BN's independently known mitigating effect (Section \ref{sec:confounds}).

\item \textbf{Dynamic path weighting:} a time-varying $D_{\mathrm{eff}}$ metric that tracks "active" pathways across training stages could reveal whether networks progressively rely on deeper paths as training proceeds, aligning with work on dynamic evolution through gradient descent (Radhakrishnan et al. \cite{Radhakrishnan2025Growing}).

\item \textbf{Extension to Vision Transformers (ViTs):} ViTs rely heavily on residual skip connections to stabilize self-attention; applying the $D_{\mathrm{eff}}$ framework to transformer-based architectures could test whether similar "functional depth" ceilings exist in attention-driven models as they scale to many layers (Dosovitskiy et al. \cite{dosovitskiy2021an}).
 
\item \textbf{Hardware-aware scaling:} combining $D_{\mathrm{eff}}$ with hardware-specific latency metrics could support a "scaling efficiency score" for edge deployment, where the trade-off between architectural depth and inference speed is critical (Ghani et al. \cite{Ghani2024Optimised}).

\item \textbf{Information-theoretic $D_{\mathrm{eff}}$:} investigating the link between effective depth and information bottlenecks, e.g., via total correlation (Xu and Shao \cite{Xu2022Information}), could help determine the minimum depth required for specific task complexities.

\item \textbf{Cross-domain validation:} although this study focuses on image recognition, the underlying principles may extend to Natural Language Processing (Supriyono et al. \cite{supriyono2024advancements}) and Graph Neural Networks (Wu et al. \cite{wu2020comprehensive}); testing $D_{\mathrm{eff}}$ on non-Euclidean data structures could clarify whether it generalizes beyond convolutional vision architectures.

\end{itemize}

\noindent Taken together, these directions would test whether the transition from nominal to effective depth analysis, demonstrated here on a bounded set of legacy CNN families under a single dataset, generalizes into a broadly useful design principle for hardware-efficient, information-theoretically grounded architectures — a claim this paper deliberately stops short of making on its own.



\appendix
\section{Effective Depth Approximation}
\label{app:effective_depth}

\setcounter{equation}{0}
\setcounter{table}{0} 
\renewcommand{\theequation}{A.\arabic{equation}}

\noindent Unlike nominal depth, effective depth is an operational measure of how information traverses transformation steps during training or inference. Following the interpretation of deep networks as ensembles of shallow paths (Veit et al. \cite{veit2016residual}), we define effective depth as the expected length of feasible input-output paths enabled by the network topology. 

Let $\mathcal{P}$ denote the set of all forward paths from input to output, and let $\ell(p)$ be the number of convolutional transformations along the path $p \in \mathcal{P}$. Under a \textit{path-uniform approximation},  \textbf{effective depth} is:

\begin{equation}
D_{\mathrm{eff}}
=
\frac{1}{|\mathcal{P}|}
\sum_{p \in \mathcal{P}} \ell(p).
\label{eq:app:effective_depth_general}
\end{equation}

\vspace{0.5cm}
\noindent \textbf{Architectural Derivations}

\begin{enumerate}[label=(\roman*)]

\item \textbf{VGG (Plain Feedforward):} These
architectures contain a single sequential path with no shortcuts or branches, so 

\begin{equation}
\mathcal{P}_{\mathrm{VGG}} = \{p\}
\end{equation}
and 
\begin{equation}
\ell(p) = D_{\mathrm{nom}},
\end{equation}

giving
\begin{equation}
D_{\mathrm{eff}}^{\mathrm{VGG}}
=
D_{\mathrm{nom}}.
\label{eq:app:effective_depth_vgg}
\end{equation}

\item \textbf{ResNets (Residual Networks):}  Residual blocks admit an exponential number of paths due to identity shortcuts. Let $\ell_{\min}$ and $\ell_{\max}$ denote the minimum path (skipping all residual branches) and the maximum path (traversing all branches), respectively. Assuming that all paths contribute equally to the ensemble (Veit et al. \cite{veit2016residual}), the effective depth is approximated as the arithmetic mean of the path limits:

\begin{equation}
D_{\mathrm{eff}}^{\mathrm{ResNet}}
\;\approx\;
\frac{\ell_{\min} + \ell_{\max}}{2}.
\label{eq:app:effective_depth_resnet}
\end{equation}

\item \textbf{GoogLeNet (Inception):} Each Inception module $m$ contains $B$ parallel branches with depths {$\{d_{m,1},\dots,d_{m,B}\}$}. The effective depth per module is the average branch depth: 

\begin{equation}
D_{\mathrm{eff}}^{(m)}
=
\frac{1}{B}
\sum_{b=1}^{B} d_{m,b}.
\end{equation}

and the total effective depth sums across the $M$ sequential Inception modules yields:

\begin{equation}
D_{\mathrm{eff}}^{\mathrm{GoogLeNet}}
=
\sum_{m=1}^{M} D_{\mathrm{eff}}^{(m)}.
\label{eq:app:effective_depth_googlenet}
\end{equation}
\end{enumerate}

\section{Gradient-Weighted Effective Depth}
\label{app:gradient_effective_depth}

\setcounter{equation}{0}
\renewcommand{\theequation}{B.\arabic{equation}}

\noindent Although ~\ref{app:effective_depth} provides a structural proxy, the \textbf{gradient-weighted effective depth}  $D_{\mathrm{eff}}^{\nabla}$ accounts for the actual propagation of the signal during backpropagation. We define the weight of a path $w(p)$  based on the $L_2$  norm of the gradients $g(p)$ flowing through it:

\begin{equation}
D_{\mathrm{eff}}^{\nabla}
= \sum_{p \in \mathcal{P}} w(p)\,\ell(p),
\label{eq:app:gradient_effective_depth}
\end{equation}
\vspace{0.5em}

\noindent
where the normalized path weights are given by:
\begin{equation}
w(p)
=
\frac{g(p)}{\sum_{p' \in \mathcal{P}} g(p')}.
\end{equation}
\vspace{0.5em}

\noindent
Paths carrying stronger gradients during training contribute more strongly to $D_{\mathrm{eff}}^{\nabla}$. Because gradients attenuate somewhat with increasing sequential depth even within a shortcut-enabled block, shorter paths tend to dominate the weighted average in residual and multi-branch architectures, producing an effective depth at or below the structural proxy.

Table B.1 reports the computed values $D_{\mathrm{eff}}^{\nabla}$, averaged over the final 20 epochs of training and over the five seeds, for the four models with more than one forward path. As reported in Section \ref{sec: validating}, the two metrics strongly correlate (Pearson r = 0.998, p = 0.002 at these four points) with a fitted slope of 0.83, indicating that the structural proxy systematically overestimates $D_{\mathrm{eff}}^{\nabla}$ by a roughly constant proportional offset (12–17\%) rather than diverging unpredictably. This supports the use of structural training-free $D_{\mathrm{eff}}$ as the primary metric in Sections \ref{sec:accuracy} -- \ref{sec: statistical_association}, while treating the offset itself and its behavior at greater depth or width, as an open question (Section \ref{sec:limitations}).


\begin{table}[t]
\caption{\textbf{Structural effective depth versus gradient-weighted effective depth. VGG models are omitted because their single-path topology makes $D_{\mathrm{eff}}^{\nabla}$ identical to the structural ($D_{\mathrm{eff}}$) by construction.}}

\label{tab:ratio}

\centering
\vspace{0.5cm}
\begin{threeparttable}
\setlength{\tabcolsep}{5.7pt}
\renewcommand{\arraystretch}{1.25}

\begin{tabular}{llcccc}
\toprule
Model & $D_{\mathrm{eff}}$  & $D_{\mathrm{eff}}^{\nabla}$ &  ($D_{\mathrm{eff}}^{\nabla}$/$D_{\mathrm{eff}}$)\\

\midrule

ResNet-18 & 10.0 & 8.3  & 0.830 \\
ResNet-34 & 18.0 & 15.1 & 0.839 \\
ResNet-50 & 26.0 & 21.8 & 0.388\\
GoogLeNet & 14.3 & 12.6 & 0.881 \\

\bottomrule
\end{tabular}
\end{threeparttable}
\end{table}


\section{Implications for NAS Heuristics}
\label{app:NAS_heuristics}

\noindent The $D_{\mathrm{eff}}^{\nabla}$ metric could, in principle, serve as a high-fidelity, zero-cost filter for the search spaces used in the recent NAS literature. By enforcing a $D_{\mathrm{eff}}^{\nabla}$ threshold during a "supernet" training phase — as suggested by the differentiable approaches in Guo et al. \cite{guo2022diff} — researchers could potentially prune "shattered"-gradient candidates before committing to full-scale training. Lightweight search regimes, such as those proposed by Lin et al. \cite{lin2024lightweight}, could, in principle, provide a more informative cost-benefit signal than parameter counts alone by identifying modules that contribute to $D_{\mathrm{nom}}$ without correspondingly reducing $D_{\mathrm{eff}}$. We emphasize that this application has not been implemented or tested in the present study; it is presented as a motivated direction for future work (Section \ref{sec:future}), not as a validated result of this paper. 

\vspace{0.5cm}






\bibliographystyle{elsarticle-num}
\bibliography{references}

\end{document}